\documentclass[final,3p,times]{elsarticle}

\usepackage{amssymb}
\usepackage{multirow}
\usepackage{graphicx}
\usepackage{times}
\usepackage{helvet}
\usepackage{courier}
\usepackage{url}
\usepackage{graphicx}  
\usepackage{xspace}
\usepackage{float}
\usepackage{multirow}
\usepackage{graphicx}
\usepackage{latexsym}
\usepackage{booktabs}
\usepackage{caption}
\usepackage{soul, color}
\usepackage{xcolor}
\usepackage{lipsum}
\usepackage{framed}
\usepackage{enumitem}

\usepackage[latin1]{}
\UseRawInputEncoding
\usepackage{amsmath, amsthm, amssymb, amsfonts}
\usepackage{centernot}
\usepackage{color}

\journal{Intelligent Based Medicine}

\begin{document}
\begin{frontmatter}

\title{\textit{VacSIM}: Learning Effective Strategies for COVID-19 Vaccine Distribution using Reinforcement Learning}

\author[1]{Raghav Awasthi}
\author[2]{Keerat Kaur Guliani} 
\author[1]{Saif Ahmad Khan}
\author[5]{Aniket Vashishtha}
\author[1]{Mehrab Singh Gill}
\author[3]{Arshita Bhatt}
\author[4]{Aditya Nagori}
\author[1]{Aniket Gupta}
\author[1]{Ponnurangam Kumaraguru}
\author[1,6]{Tavpritesh Sethi\corref{correspondingauthor}}
\cortext[correspondingauthor]{Corresponding author}
\address[6]{Office: A-309, (R \& D Block),Indraprastha Institute of Information Technology Delhi,110020}
\ead{tavpriteshsethi@iiitd.ac.in}
\fntext[1]{Indraprastha Institute of Information Technology Delhi}
\fntext[2]{Indian Institute of Technology Roorkee}
\fntext[3]{Bhagwan Parshuram Institute of Technology, New Delhi}
\fntext[4]{CSIR-Institute of Genomics and Integrative Biology, New Delhi}
\fntext[5]{Maharaja Surajmal Institute of Technology, New Delhi}

\begin{abstract}

  A \textit{COVID-19} vaccine is our best bet for mitigating the ongoing onslaught of the pandemic. However, vaccine is also expected to be a limited resource. An optimal allocation strategy, especially in countries with access inequities and temporal separation of hot-spots, might be an effective way of halting the disease spread. We approach this problem by proposing a novel pipeline \textit{VacSIM} that dovetails Deep Reinforcement Learning models into a Contextual Bandits approach for optimizing the distribution of \textit{COVID-19} vaccine. Whereas the Reinforcement Learning models suggest better actions and rewards, Contextual Bandits allow online modifications that may need to be implemented on a day-to-day basis in the real world scenario. We evaluate this framework against a naive allocation approach of distributing vaccine proportional to the incidence of \textit{COVID-19} cases in five different States across India (Assam, Delhi, Jharkhand, Maharashtra and Nagaland) and demonstrate up to 9039  potential infections prevented
 and a significant increase in the efficacy of limiting the spread over a period of 45 days through the \textit{VacSIM} approach. Our models and the platform are extensible to all states of India and potentially across the globe. We also propose novel evaluation strategies including standard compartmental model-based projections and a causality-preserving evaluation of our model. Since all models carry assumptions that may need to be tested in various contexts, we open source our model \textit{VacSIM} and contribute a new reinforcement learning environment compatible with OpenAI gym to make it extensible for real-world applications across the globe.  \footnote{http://vacsim.tavlab.iiitd.edu.in:8000/} 
 
\end{abstract}

\begin{keyword}  COVID-19, Vaccine Distribution, Policy Modeling, Reinforcement Learning, Contextual Bandits Problem\\
\end{keyword}
\end{frontmatter}
\section{Introduction}
Vaccines have played a crucial role in combating infectious diseases for hundreds of years\cite{greenwood_contribution_2014}. The successful eradication of smallpox, a transmittable disease responsible for causing huge casualties, happened due to the gradual and effective widespread use of vaccines\cite{noauthor_smallpox_nodate}. Avoidance of millions of deaths and side effects caused by diseases like polio and tetanus was because of vaccination efforts which prove their impact and significance for global health. Despite this impact, if such a vital resource is not optimally distributed and allocated to the regions and communities where its need is the most, this can lead to huge repercussions that could aggravate the situation and can increase the risk of escape variants in mixed and partially vaccinated communities \cite{rella_rates_2021}.  
All countries across the globe have eagerly waited for the launch of an effective vaccine against \textit{SARS-CoV-2}. In India, the vaccine allocation process started on a first-come-first-serve basis, from January 2021 with the launch of the CoWIN \footnote{https://www.cowin.gov.in/} portal by the Government of India. As potential candidates continue to introduce their products in the market, there is an urgent need for optimal distribution strategies that would mitigate the pandemic at the fastest rate possible\cite{white_proposed_2020}\cite{khamsi_if_2020}. At the current rate of vaccination estimated to be 2.2 million doses per day, India may only be able to cover 30\% of its population by end of this year, triggering concerns about a third wave hitting India, especially its rural parts, in the absence of herd immunity\footnote{https://api.covid19india.org/}. 
Here we summarize the key factors that will need to be considered for effective mitigation:

\begin{itemize}
     \item \textbf{Scarcity of supply:} Despite large scale production efforts, it is expected that the vaccine will still be a scarce resource as compared to the number of people who would need it. In addition to the vaccine itself, there may also be scarcity in the components leading to its delivery, e.g. syringes. Many researchers estimated that 60-70\% of the  population would require vaccination to achieve herd immunity. This implies that India would need 1.89 billion syringes to deliver the vaccine en-masse. \footnote{https://www.nature.com/articles/d41586-021-00728-2}. 
 
 \item \textbf{Need for Equitable distribution and Digital inclusion:} A truly equitable distribution will not just be defined by the population or incidence of new cases alone, although these will be strong factors. Other factors ensuring equity of distribution include quantum of exposure, e.g. to the healthcare workforce that needs to be protected. In this paper, we assume that the exposure is proportional to the number of cases itself, although the proposed methodology allows more nuanced models to be constructed. There may also be unseen factors such as vaccine hoarding and political influence, which are not accounted for in this work. Another issue that has recently come into light is regarding the increased inaccessibility of vaccines in India, which are largely booked through the online CoWIN portal, owing to the significant digital divide in the country's population. 
 \item \textbf{Transparent, measurable and effective policy:} The design of policy would need to be guided by data, before, during and after the vaccine administration to a section of the population. Since the viral dynamics are rapidly evolving, the policy should allow changes to be transparent and effects to be measurable in order to ensure maximum efficacy of the scarce vaccine resource. On the larger scale of states and nations, this would imply continuous monitoring of incidence rates vis-a-vis a policy action undertaken. 

\end{itemize}

Although the aforementioned factors seem straightforward, the resulting dynamics that have emerged during the actual roll-out of the vaccine are far too complex for optimal decision making by humans. The daunting nature of such decision-making can be easily imagined for large and diverse countries such as India, especially where health is a state subject. Artificial intelligence for learning data-driven policies is expected to aid such decision making as there would be limited means to identify optimal actions in the real world. A ‘near real-time’ evaluation as per the demographic layout of states and consequent initiation of a rapid response to contain the spread of COVID-19\cite{foster_survey_nodate} will be required. Furthermore, these policies will need to be contextualized to the many variables governing demand or ‘need’ for the vaccine distribution to be fair and equitable\cite{deo_covid-19_2020}. Ground testing of these scenarios is not an option, and countries will have to face this challenge. \\
Reinforcement Learning (RL) algorithms have shown significant progress for decision making in medicine and public health \cite{liu_reinforcement_2020} \cite{saria_individualized_2018} \cite{raghu_continuous_nodate} \cite{komorowski_artificial_2018}.
In this paper, we introduce \textit{VacSIM}, a novel feed-forward reinforcement learning approach for learning effective policy combined with near real-time optimization of vaccine distribution and demonstrate its potential benefit through the example of five states presenting contrasting contexts in India. Since real world experimentation was out of the question, the change in projected cases obtained via a standard epidemiological model was used to compare the \textit{VacSIM} policy with a naive approach of incidence-based allocation. Finally, our novel model is open-sourced and can be easily deployed by policymakers and researchers, thus can be used in any part of the world, by anyone, to make the process of distribution more transparent.

\section{Methods:}
To get an optimal distribution of vaccine, we proposed a novel pipeline where we joined Deep Reinforcement Learning models with a Contextual Bandits model in a feed-forward way which makes the optimization process more robust, context-specific and easily deployable in real-time. In the following sections, we are adding some background information on  Reinforcement Learning Algorithms in order to provide a sense of our pipeline.

\subsection{Q Learning:}

Policy learning algorithms can be broadly categorized into two types: off-policy and on-policy. Off-policy learning algorithms are preferred as these allow the agent to learn the optimal policy regardless of the agent's actions (under the current policy). This not only facilitates faster learning (convergence) but also avoids learning of a sub-optimal policy by the agent as it can explore continuously. One of the most popular off-policy algorithms in \textit{RL} is Q-learning. Q-Learning uses the Bellman Equation for iterative updating: 

\large\begin{equation} {\label{eq1}
Q^*(s,a)=E[r+\gamma max_{a'} Q^*(s',a')]}
\end{equation}
\normalsize

where \(r\) is the immediate reward and \(\gamma\) is the discount factor.
Given that an agent follows a policy \(\pi\), then Q$_\pi$(s,a) denotes the expected return or reward obtained by the agent if it chooses action $a$ at state $s$ and then keeps on following the policy \(\pi\). $Q(s,a)$ estimates how favourable it is to choose action $a$ at state $s$ according to the current policy. The optimal Q function, $Q^*(s, a)$ denotes the maximum return or reward that can be achieved by first choosing action $a$ at state $s$ and then following the optimal policy. As evident from equation (1), $Q^*(s, a)$ is the sum of the immediate reward and the maximum reward possible thereafter from state $s^{\prime}$ discounted by \(\gamma\) (same as \(\gamma\) discounted maximum Q value for state $s^{\prime}$). However, Q-Learning becomes impractical for environments where there may be a large (potentially infinite) set of actions to choose from. An example of a large environment with continuous action space \cite{komorowski_artificial_2018} is optimization of number or syringes or medication dosage.

\subsection{Deep Q-Network:}
Deep Q-Networks \textit{(DQNs)} are useful in the presence of large environments with elaborate state and action spaces. Here Q-learning is extended to \textit{DQN} by using a deep neural network used to map input states to (action, Q-value) pairs\cite{hester_deep_2018}\cite{van_hasselt_deep_2015}. Most real-world scenarios such as  healthcare resource allocation have complex state-action spaces, and thus need advanced approaches such as \textit{DQNs}. A  \textit{DQN} minimizes the loss at each iteration $i$ as given below:

\large
\begin{equation} \label{eq2}
L_i(\theta_i) = E_{s,a,r,s'\sim~\rho(.)}[((y_i - Q(s,a;\theta_i))^2]
\end{equation} 
\normalsize

where y$_i$ = $r$ + \(\gamma\) $max$$_{a'}$ %
$Q$($s^{\prime}$,$a^{\prime}$;\(\theta\)$_{i-1}$) is the Temporal Difference (TD) target; y$_i$ - Q is the TD error and \(\rho\) represents behaviour distribution.

In our work, we employ \textit{DQN} to better parameterize real-world conditions and make our model more efficient in terms of time and memory requirements.

\subsection{Actor-Critic using Kronecker-Factored Trust Region (ACKTR):}

Q-learning and DQN are both value-based algorithms, where the goal is to learn a single deterministic action from a discrete set of actions by finding the maximum Q value. There are no trainable parameters in value-based methods that control the probabilities of action. Thus, a value-based method cannot solve an environment where the optimal policy is stochastic requiring specific probabilities. To overcome this challenge, we need to learn a stochastic map from state to action. In Policy Gradients \cite{williams_simple_1992} \cite{schulman_high-dimensional_2018}, we usually use a neural network (or other function approximators) to directly model the action probabilities $\pi_\theta(a|s)$. Each time the agent interacts with the environment, we tweak the parameters \(\theta\) of the neural network so that “good” actions (actions giving maximum reward)  will be sampled more likely in the future. We repeat this process until the policy network converges to the optimal policy \(\pi^*\). 
\\ Formally, the objective of Policy Gradients is to maximize the total future expected rewards \(E[R_t]\), where \(R_t\) represents the sum of future discounted rewards. Whereas, in  Q-learning and DQN the “policy” followed the action that maximized the Q value at each step, actor-critic methods combine Policy Gradient methods with a learned value function. With actor-critic methods \cite{mnih_asynchronous_2016}, we learn two different functions: the policy (or “actor”), and the value (or the “critic”). 
In this paper, we used the ACKTR algorithm (on-policy algorithm) \cite{wu_scalable_2017}  with the Policy Gradient method which follows an actor-critic architecture: one deep neural network to estimate the policy and another network to estimate the advantage function (a measure of how good or bad a certain action is at a given state). ACKTR uses the Kronecker-factored approximation \cite{ly_tutorial_2017} \cite{martens_new_2020} to optimize both the actor and critic.

\subsection{Bandit Algorithms: Multi-armed bandits and Contextual Bandits problem}
 Reinforcement algorithms  being computationally expensive and time-consuming, cannot be used in real-time decision making. Contextual Bandits \cite{collier_deep_2018} algorithms on the other hand, are space and time-efficient. These are extension of multi-armed bandits algorithms. Multi-armed bandits can be simply be understood as a gambler with a fixed amount of money taking a sequence of actions on multiple slot machines while trying to optimize the overall reward. The $K$ machines have reward probabilities, ${\theta_1,…,\theta_K}$. ⟨Action($A$),Reward($R$)⟩ tuples describe a multi-armed bandit, where at each time step \textit{t}, an agent takes an action $a$ on one slot machine and receives a reward $r$. The goal is to maximize the cumulative reward. 
 This also corresponds to minimization of the potential loss by not picking the sub-optimal action, in case optimal actions with best reward are known, e.g. vaccination of a high-risk individual.
The optimal reward probability $\theta^\ast$ of the optimal action $a^\ast$ is:

\large
\begin{equation}  \label{eq7}
    \theta^\ast=Q(a_\ast)=\underset{a\in S}{\max}\{Q(a)\}=\underset{1<i<K }{\max}\{\theta_i\}
\end{equation}
\normalsize

where $Q(a)$ is the expected reward.

The Loss function or \textit{Regret}\cite{riquelme_deep_2018} is a conceptual function to understand and optimize the performance of the Bandits problem. Since we don’t know if an action played was the most ‘reward-fetching’, rewards against all the actions that can be played are sampled, and the difference between the action chosen and the action against the maximum reward is defined as ‘regret’. Therefore, minimizing regret achieves the goal of maximizing reward. The total regret we might have by not selecting the optimal action up to the time step \textit{T}:

\large
\begin{equation}  \label{eq8}
    L_t = E[\sum_{t=1}^{t=T}(\theta^\ast - Q(a)]
\end{equation}
\normalsize

Thus, a multi-armed bandit algorithm returns the best set of actions, provided the context is fixed. 
However, a multi-armed bandit algorithm does not account information from the changing context, a common occurrence in real-world scenarios such as the evolving pandemic. Therefore the agents may end up playing the same action multiple times even though the context may have changed, thus getting stuck at a sub-optimal condition. To circumvent this problem, Contextual Bandits algorithm is proposed which  contextualizes the choice of the bandit to its current environment. Contextual Bandits, represented as a tuple ⟨Action($A$),Context($C$),Reward($R$)⟩, play an action based on its current context, given a corresponding reward, hence are more relevant to real world environments such as the vaccine distribution problem addressed in this work. Given, for time $t$ = 1...n, a set of contexts \(C\) and a set of possible actions \(A\) and reward/payoffs \(R\) are defined. At a particular instant, based on a context \(c_t \in C\), an action \(a_t \in A\) is chosen and a reward or a payoff

\large\begin{equation}  \label{eq9}
    R = E[r \mid c, a]
\end{equation}\normalsize
is obtained.  For an optimal action \(a^* \in A\) such that the expectation of reward against this action is maximum, 
\large\begin{equation}  \label{eq10}
    r^* = \underset{{a_t}\in A}{\max}{ E(r \mid c, a_t)}
\end{equation} \normalsize
    
\hspace{-0.5cm}the regret can be expressed as 
\large\begin{equation}  \label{eq11}
    Z = [r^* - E(r \mid c, a_t) ]
\end{equation}\normalsize
and \textit{cumulative regret} can be expressed as 
\large\begin{equation}  \label{eq11}
    Z^* =  \sum_{t=1}^n Z
\end{equation} \normalsize

\hspace{-0.55cm}Contextual Bandits model was implemented using the python package of Vowpal Wabbit \cite{noauthor_vowpal_nodate}.

\subsection{\textit{VacSIM}: A Feed forward Approach}
In \textbf{VacSIM}, we concatenated the Deep Reinforcement Learning models (i.e. ACKTR or DQN) and  Contextual Bandits model in a feed-forward manner to create an off-policy learning framework. This was incorporated to select the optimal policy through the Contextual Bandits approach from the ones generated by the Deep Reinforcement Learning models, as shown in \textbf{Figure 1}. This was done to address the following challenges that need to be tackled in real world problems such as optimal vaccine distribution:
\begin{itemize}
    \item \textbf{Solving in real-time:} A vaccine distribution problem is expected to be fast-paced. Thus, an overwhelming amount of brainstorming with constrained resources and time would be required to develop an effective policy for the near future. This calls for the development of a prompt and an easily reproducible setup.  
    \item \textbf{Lack of ground truth:} This is one of the key challenges in this paper. Since the roll-out of the vaccine will not give us the liberty of testing various hypotheses, a lack of ground truth data is generally expected. This is analogous to zero-shot learning problems, thus precluding a simple supervised learning-based approach.
    \item \textbf{Absence of evaluation with certainty:} Lack of ground testing naturally implies nil on-ground evaluation. In that case, it often becomes challenging to employ evaluation metrics that offer a significant amount of confidence in results. In order to solve this problem, we rely upon the simulated decrease in the number of susceptible people as the vaccine is administered.
    \item \textbf{Model scaling:} We have ensured that the learning process of the models simulates the relationship between different objects in the real world environment accurately, and at the same time can be scaled down in response to computational efficiency and resource utilization challenges. This is done by choosing the right set of assumptions that reflect the real world scenario. 

\end{itemize}

\begin{figure*}[h]
\begin{center}

\includegraphics[width=16cm]{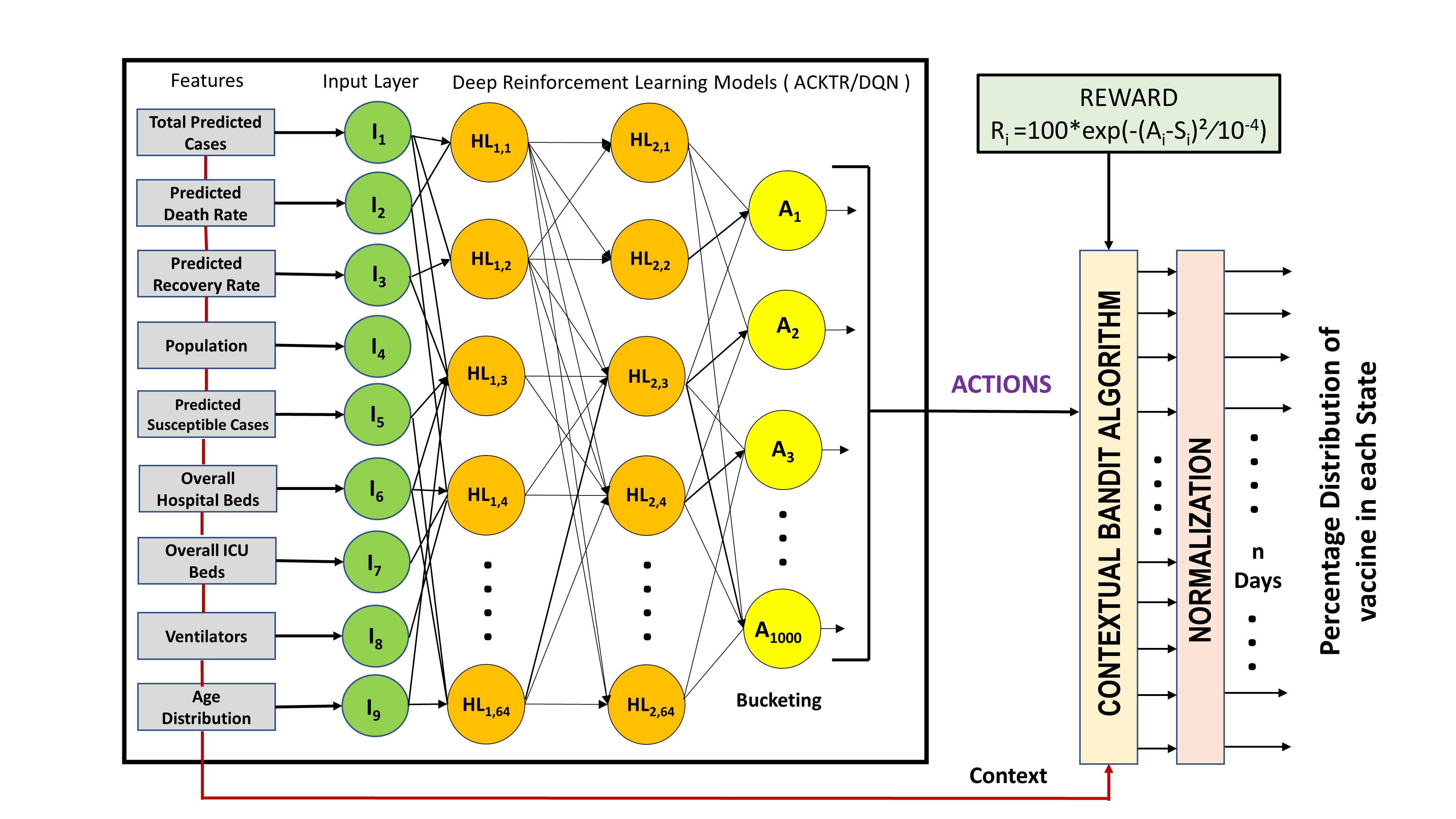} \\
    \textbf{Figure 1: {{VacSIM} architecture: A novel feed forwarded pipeline for policy learning of COVID-19 vaccine optimal distribution. The actions and rewards obtained from the Deep Reinforcement Learning models were fed forward into the training of Contextual Bandit algorithm so that faster optimal online decisions could be calculated for further dates taking in context, the ever changing demographics of the states. }}
\end{center}
\end{figure*}

We pipelined the Deep Reinforcement Learning models with a Contextual Bandits approach where recommendations for vaccine distribution policy(generated by the RL models) were used as training data for the Contextual Bandits. These models were trained separately using the same reward function defined in equation (6).
 \\Reinforcement Learning approaches replicate human decision-making processes. However, the absence of evaluation makes them less trustworthy, especially when real lives may be at stake. To this end, we also presented two different novel model evaluation techniques in this paper:

 \begin{enumerate}
    \item Estimating future impact of vaccination by projecting COVID-19 cases, using SEIR models.\cite{li_global_1995}
   \item { Learning Bayesian Network to verify the underlying structure of COVID-19 trajectory indicators}.
     
     \end{enumerate}

\section{Results}

\begin{figure*}[hbt!]
\begin{center}
\includegraphics[width=16cm]{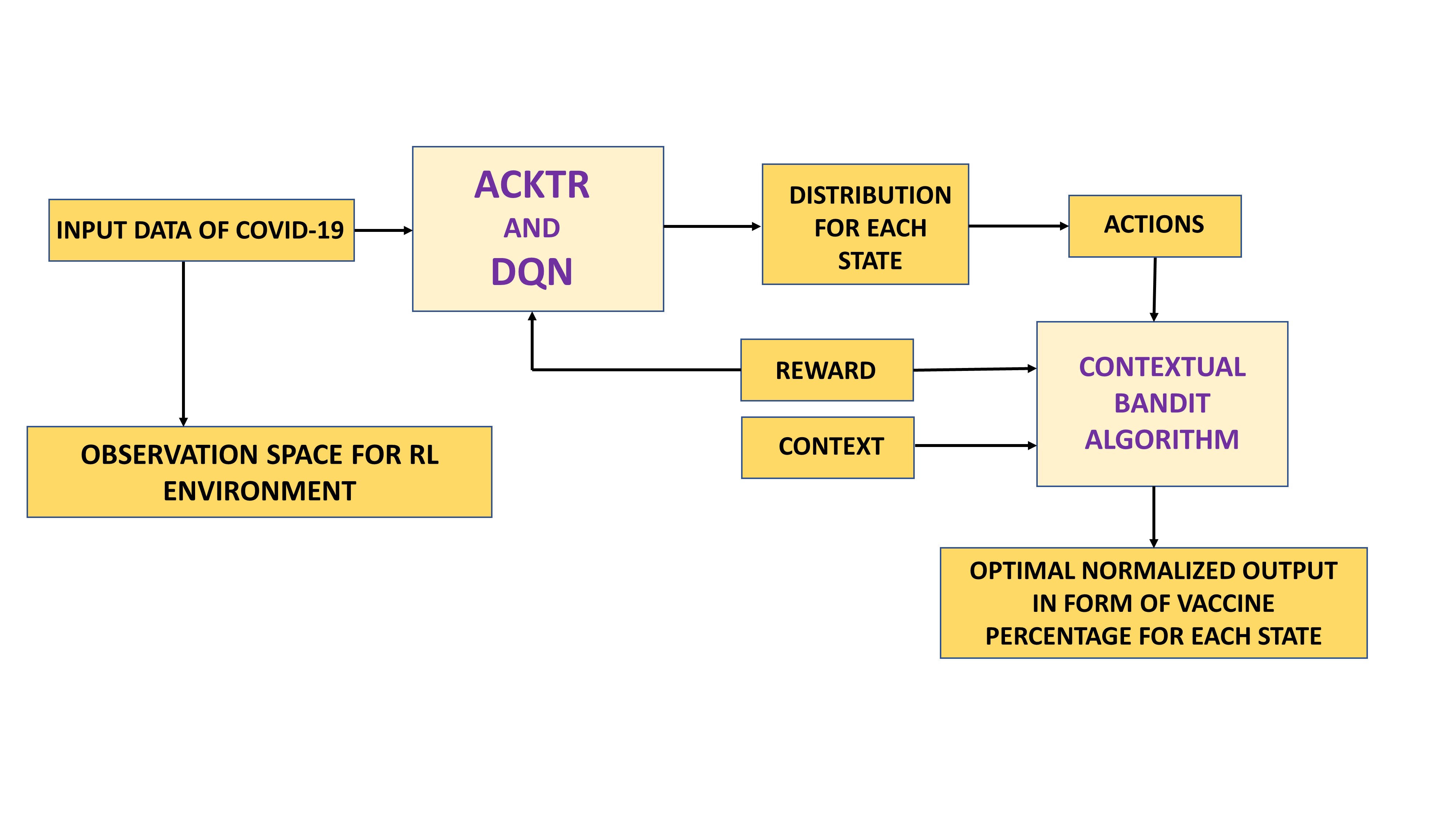}

 \textbf{Figure 2: {Flexible model setup for optimization of vaccine distribution policy in India using the {VacSIM} approach where components can be easily replaced with alternatives and adopted for diverse settings.}}
\end{center}
\end{figure*}

\begin{table}[]
\centering
\begin{tabular}{@{}lll@{}}
\textbf{S.No.} & \textbf{Feature} & \textbf{Description} \\ \midrule
1 & Predicted Death Rate & The percentage ratio of the predicted deaths in the State to the total predicted \\ && cases in that State calculated using projections obtained from a fitted standard \\ && \textit{Compartmental model}, i.e. a Susceptible, Exposed, Infected and Recovered \\ && (\textit{SEIR}) model.\\
2 & Predicted Recovery Rate & The percentage ratio of the predicted recoveries in the State to the total \\ && predicted cases in that State using projections obtained from the SEIR model.\\
3 & Population of a State & We extracted the population for each State from the 2011 census data \\ && conducted by the government of India\cite{noauthor_census_nodate}. \\
4 & Predicted Susceptible Cases of a State & We estimated the susceptible population as the difference between the \\ && population of a particular State and the total number of predicted infected \\ && cases of that State. \\
5 & Hospital Facilities in a State & We used overall Hospital Beds, overall ICU Beds and Ventilators data in \\ && our models\cite{noauthor_development_nodate}. \\
6 & Age Distribution of a State & In order to prioritize vulnerable population, we considered people with age \\ && greater than 50.\\
\bottomrule
\end{tabular}
\captionsetup{font=normalsize}
\caption*{\textbf{Table 1: Features present in context space used in our models}}
\end{table}

\subsection{Model Setup}
We extracted the State-wise time series data of COVID-19 characteristics from the website of Ministry of Health and Family Welfare of India. \footnote[1]{https://mohfw.gov.in/} and used them in the experiments as described in \textbf{Figure 2}. In order to determine vaccine allocation in the near future, we require future scenarios of COVID-19. Thus, in order to train VacSIM, we projected Susceptible, Confirmed, Recovered cases from 1 December 2020  to 26 December 2020 using SEIR model (refer supplementary for model details). The five States chosen for this study, i.e., Assam, Delhi, Jharkhand, Maharashtra and Nagaland are representative of possible scenarios for the vaccine distribution problem, i.e., high incidence (Maharashtra and Delhi), moderate incidence (Assam) and low incidence (Jharkhand and Nagaland). In choosing the five different States, we hope to generalize our predictions to other States across the spectrum while minimizing the bias introduced into the learning by a widely variant \textit{COVID-19} incidence across the country.

\noindent In order to avail vaccine for the highly vulnerable population and to combat COVID-19 infection spread, we selected the context comprising of features listed in \textbf{Table 1}.

\noindent The implementation of \textit{VacSIM} pipeline is detailed henceforth.

\subsection{Deep Reinforcement Learning Models:} Open-AI\cite{openai_gym_nodate} stable-baselines framework was used to construct a novel and relevant environment suited to our problem statement for ACKTR and DQN to learn in this environment.

 \subsubsection{Input} We considered the observation space($s$) as a grid over context vectors with the latter describing the context vector of states. These include   Population, Age Distribution, Ventilators, Overall ICU Beds, Overall Hospital Beds, Predicted Susceptible Cases, Total Predicted Cases, Predicted Death Rate, and Predicted Recovery Rate at a given time. The predicted rates were estimated by using standard SEIR projections. \\
   The Action space describes the quantum of vaccines distributed to each of the states across India, represented as discrete buckets. For example, a bucket size of 1000 implies a batch of 1000 vaccines rolled out to a particular state of India.  Formally, the action space $a$ was defined as,
    $a$= $\{b_0,b_1,b_2,b_3,...,b_i,...,b_{n-1} \}$
    
   \large\begin{equation}
   b_i = i*\frac{Total \; Available \; Vaccine}{n}
   \end{equation}
   \normalsize
where \textit{n} is defined as \textit{bucket size} and considered as a hyper parameter.

\subsubsection{Training} Following are the assumptions used while building the environment:
    \begin{itemize}
        \item The nature of the vaccine is purely preventative and not curative, i.e.,
        it introduces immunity when administered to a susceptible person against future COVID infection but plays no role in fighting against existing infections (if any).
        \item The vaccine has 100\% efficacy, i.e. all people who are vaccinated will necessarily develop immunity in 45 days\cite{polack_safety_2020} against future \textit{SARS-CoV-2} infection. This assumption is easily modifiable and is expected to have little consequence in deciding the optimal allocation unless data reveal differential immunity response of certain populations within India. However, we leave scope for it to be modified as per the situation by taking this as a hyper-parameter for \textit{VacSIM}. 
    \end{itemize}
    \textbf{Reward Function:}
    The reward function was designed to maximize the decrease in the susceptible population count with the minimum amount of vaccine, considering it to be a limited resource. 
    
\large\begin{equation} \label{eq12}
R_i = 100*exp(-(A_i-S_i)^2/10^{-4})
\end{equation} \normalsize

where \(R_i\) is the reward given by the environment; \(S_i\) is the fraction of susceptible population in a particular State and \(A_i\) is the fraction of vaccine distributed to the State by the model.\\

\noindent \textbf{The Explore-Exploit Trade-off:} We tap into the explore-vs-exploit dilemma, allowing the model to reassess its approach with ample opportunities and accordingly, redirect the learning process. We set the exploration rate at 10\% for DQN. However, this too is flexible and can be treated as a hyper-parameter.
Here we reason that the local context is most strongly influenced by its own past values, hence we trained our model independently for the selected 5 States from 1 December to 26 December, in order to maintain spatial independence.10000 and 2000 iterations \textbf{Figure 3} were used to learn DQN and ACKTR respectively. Iterations for training were decided based on the stability of the reward function curve. Where Iterations refers to one complete exploration of environment by a model agent\\

\begin{figure*}[hbt!]
\begin{center}
\includegraphics[width=\textwidth]{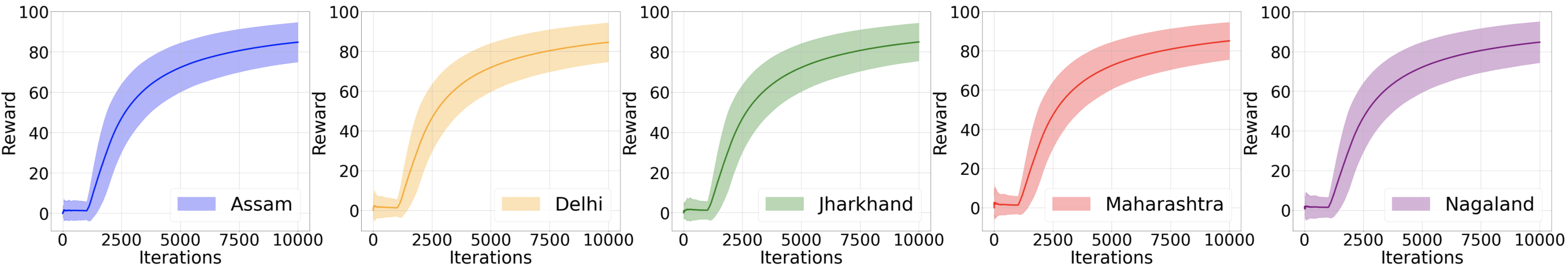}

\includegraphics[width=\textwidth]{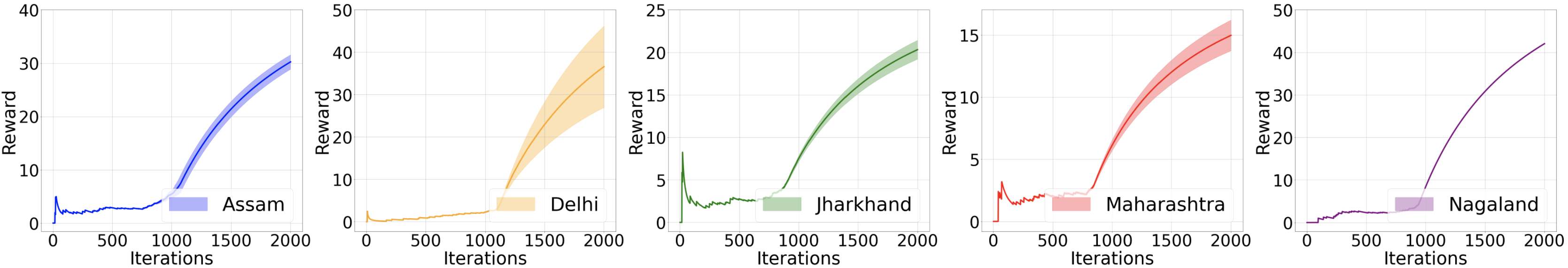}
\\
 \textbf{Figure 3: Day-wise rewards at every iteration were recorded for all the states. The Mean and SD obtained over 26 day projections are shown as confidence bands. {(Top) Smoothed and Increasing Reward Curve of DQN (Bottom) Smoothed and Increasing Reward Curve of ACKTR.}}
\end{center}
\end{figure*}

\noindent \textbf{Hyper-parameters:} A complete list of the hyper-parameters is given in \textbf{Table 2}.

\begin{table}[hbt!]
\centering
\begin{tabular}{@{}lll@{}}
\textbf{S.No.} & \textbf{Hyperparameter Name} & \textbf{Hyperparameter Value} \\ \midrule
1 & Batch Size: number of vials in one round of distribution. & 1000000 \\
2 & Exploration Rate of DQN & 10\% \\
3 & Vaccine efficacy & 100\% \\
4 & Number of days to reach full efficacy & 45 \\
5 & Bucket size & 1000 \\
6 & Number of recipients per day & 5 \\ \bottomrule
\end{tabular}
\captionsetup{font=normalsize}

\caption*{\textbf{Table 2 :Hyper-parameters used during Policy learning}}
\end{table}

\subsubsection{Output} The output of the Deep Reinforcement Learning models  was a distribution set dictating the share of each recipient in the batch of vaccines dispatched at the time of delivery.

\subsection{Contextual Bandits Model} The output of the Deep Reinforcement Learning model is spanned over 26 days (1 December-26 December 2020). The distribution sets so obtained were scaled to bucket sizes ranging from 100 to 1000 with an increment step of 100.
The actions of the five States associated with each bucket size were normalized to get the percentage distribution ratio for all States. Normalized here refers to the percentage ratio of a given distribution set of a State for a given bucket size and the sum of the distribution sets of the five States over that bucket size for a given date. 

\subsubsection{Training} We considered 1 December-26 December data along with distribution sets and the corresponding set of rewards obtained from the Deep Reinforcement Learning models as training data for Contextual Bandits.
 
 \subsubsection{ Number of actions and features} 
 The action space and the features in the Context space are the same as those in the Deep Reinforcement Learning model.
 
 \subsubsection{Testing} Using the Context, actions and the corresponding set of rewards as the training dataset, we tested the model day-wise for a period of five days (27 December-31 December) with each day having 10 possible bucket sizes, i.e. 100 to 1000 for each State as output.

\subsubsection{Output} The unadjusted actions (which were not normalized) obtained after testing the model were first adjusted bucket-wise for each State for a particular date by taking the percentage ratio of the unadjusted action of that State and the sum of unadjusted actions of all five States.\\
We consider that the vaccine is distributed on 31 December 2020. The consequent distribution of vaccines among the five States for each bucket size was then evaluated. Our models learn well for the bucket range 200-500. The vaccine distribution for the same in case of the VacSIM policies is shown in \textbf{Table 3}. It was observed that both  ACKTR+CB and DQN+CB yielded similar results, which is conceptually expected as both were aimed at reducing the overall susceptible population as optimally as possible.

\begin{table}[hbt!]
\centering
\resizebox{0.8\textwidth}{!}{%
\begin{tabular}{lllllll}
\hline
\textit{\textbf{Model}} & \textit{\textbf{Bucket Size}} & \textit{\textbf{Assam}} & \textit{\textbf{Delhi}} & \textit{\textbf{Jharkhand}} & \textit{\textbf{Maharashtra}} & \textit{\textbf{Nagaland}} \\ \hline
\multirow{4}{*}{ACKTR + CB} & 200 & 15.5 & 8 & 17.5 & 58 & 1 \\ \cline{2-7} 
 & 300 & 15.5116 & 7.9208 & 17.4917 & 57.7558 & 1.3201 \\ \cline{2-7} 
 & 400 & 15.5172 & 8.1281 & 17.734 & 57.3892 & 1.2315 \\ \cline{2-7} 
 & 500 & 15.5206 & 8.055 & 17.6817 & 57.3674 & 1.3752 \\ \hline
\multirow{4}{*}{DQN + CB} & 200 & 15.8974 & 8.2051 & 16.9231 & 58.4615 & 0.5128 \\ \cline{2-7} 
 & 300 & 15.9864 & 8.1633 & 17.0068 & 58.1633 & 0.6803 \\ \cline{2-7} 
 & 400 & 16.1616 & 8.3333 & 17.1717 & 57.5758 & 0.7576 \\ \cline{2-7} 
 & 500 & 16.129 & 8.2661 & 17.1371 & 57.6613 & 0.8065 \\ \hline
\end{tabular}%
}
\captionsetup{font=normalsize}

\caption*{\textbf{Table 3: VacSIM model output on different bucket size} }
\end{table}

\subsection{Model Evaluation through Projection Scenario using SEIR models:}
SEIR model is a differential equation-based epidemiological model used to simulate infection future trajectory. In the SEIR model,  the total population N divides into 4 compartments S,E,I \& R. Term S
represents the susceptible individuals, E includes the latent phase of covid where the individual
is infected but not infectious. I represent infected individuals and R represents the recovered
individuals.
Since there is no way that the evaluation of distribution policy can be done in the absence of a real world distribution, we defined the Naive baseline distribution policy as $\%$ of vaccine given to a State $=$ $\%$ of Infected People in that State and compared it with our model's learned distribution. With 10,00,000 doses and 5 States, we simulated the distribution of the available vaccine on 31 December for the Naive and \textit{VacSIM} policies. The number of resulting (projected) infections for 45 days after the vaccine distribution was calculated using the SEIR Model. For each bucket size (200,300,400,500), day-wise total cases of all five States were summed up for both models (ACKTR/DQN) and then their differences were measured with total cases for all five States arising after Naive distribution as shown in Figure 4.
Our results indicate that the \textit{ACKTR} based policy additionally reduces a total of 8845, 7787, 4703 and 5103 projected infected cases, with bucket sizes 200, 300, 400 and 500 respectively, in the next 45 days. 
Likewise, \textit{DQN} based policy additionally reduces a total of 9039, 6686, 5355 and 4698 projected infected cases, with bucket sizes 200, 300, 400 and 500 respectively.

\begin{figure*}[hbt!]
\begin{center}
\includegraphics[width=13.5cm,height=13.5cm]{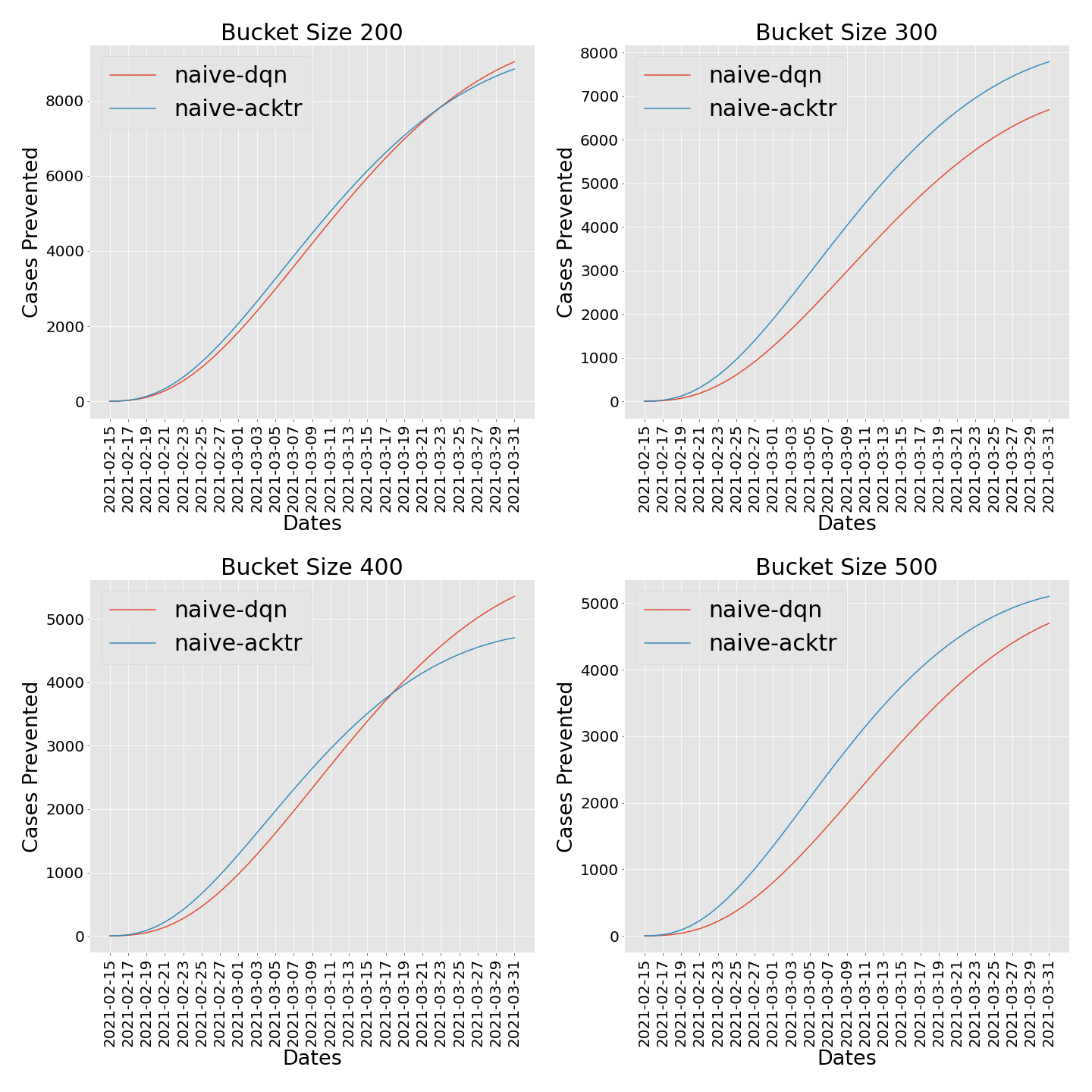}
\\
\textbf{Figure 4: {Additional projected infection cases prevented in next 45 days by following VacSIM driven approach instead of Naive approach.}}
\end{center}
\end{figure*}

\subsection{Model evaluation through learning the causal structure of simulated data obtained:}
The ultimate goal of vaccine distribution is to reduce mortality and morbidity. Since our model relies entirely upon simulations, in the absence of a real-time vaccine data, we checked if the data generated by such an approach follow the cause-and-effect relationships as expected in the real world data. 
A Bayesian network is a special class of probabilistic graphical models which is useful for modeling causal relationships. Being a graph with directed edges without cycles, the parent-child relationships can be interpreted as cause and effect relationships under strong constraints. The directed acyclic graph (DAG) $G$ is defined as a triple, $N = \{X, G, P\}$ defined that can be over a set of random variables, $X$. It encapsulates the underlying joint probability distribution $P$ that can be factored as the product of probabilities of each node $v$ conditioned upon its parents $par(v)$.

\large\begin{equation} 
\begin{split}
P(X) = \prod_{v\in V}{P(x_v | x_{pa(v)})}
\end{split}
\end{equation}
\normalsize

where v corresponds to the random variable \textit{v} and \textit{par(v) are the set of  parent variables.}

\textit{Structure-learning} of Bayesian network was carried out using \textit{Hill Climbing} algorithm\cite{gamez_learning_2011} with \textit{Akaike Information Criterion (AIC)} and \textit{Bayesian Information Criterion (BIC)} as scoring functions, and ensemble averaging over 501 bootstrapped networks. These models were learned using \textit{wiseR} package\cite{sethi_wiser_2018}.

\begin{figure*}[hbt!]
\begin{center}
\includegraphics[width=\linewidth,height=8cm]{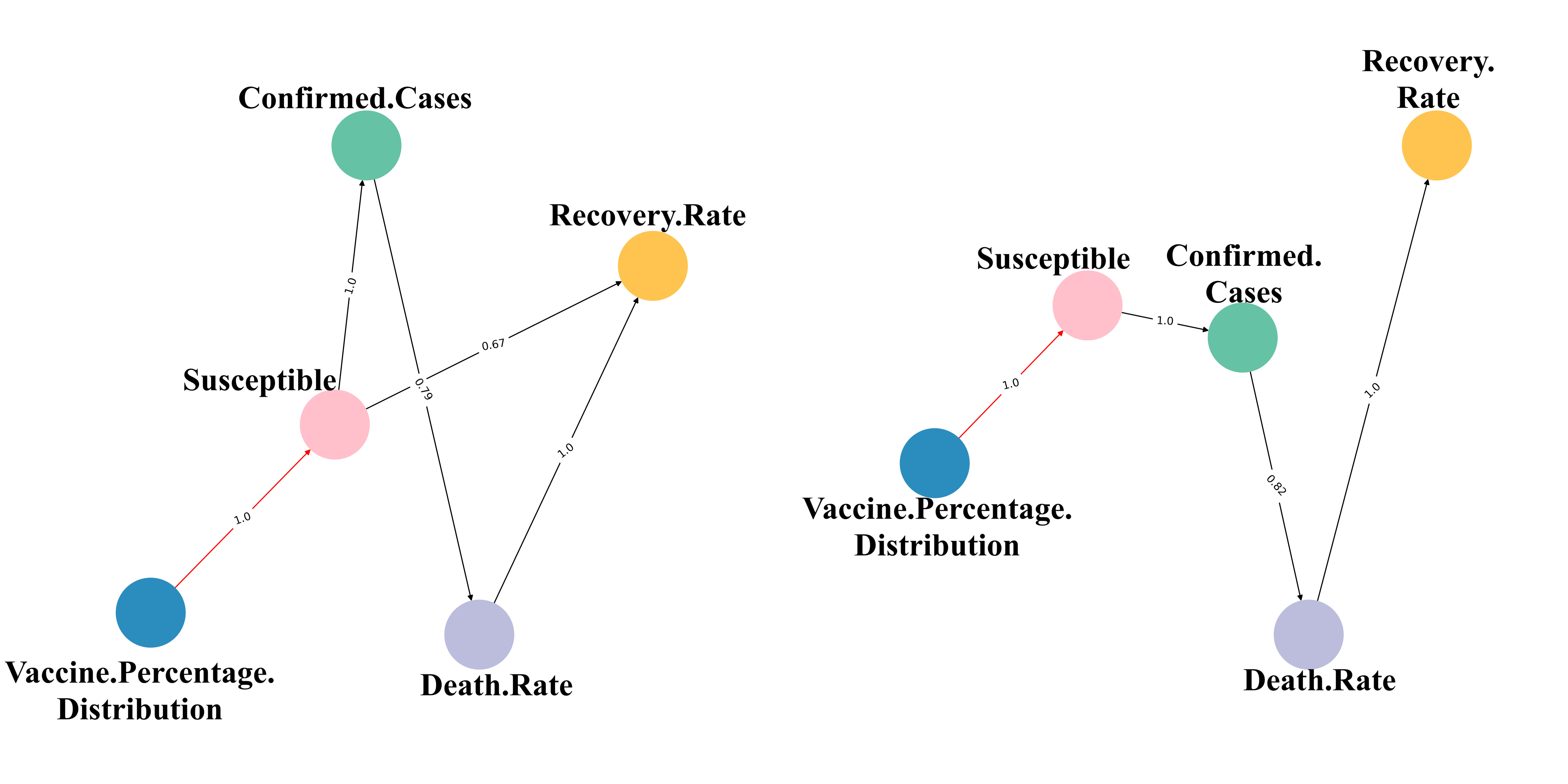}

    \textbf{Figure 5: {Ensemble averaged causal structure of the Bayesian network obtained from 501 bootstraps, using Hill Climbing optimizer for AIC (left) and BIC (right) as scoring functions. Vaccine Percentage obtained from the model was observed as a parent node of Susceptible cases, thus indicating the causality preserving nature of {VacSIM} simulations.}}
\end{center}

\end{figure*}

State-wise time series data of death, recovery, infected people, susceptible people and the amount of vaccine obtained from our model were used to learn the structure. Blacklisting among nodes was done such that vaccine percentage cannot be the child node of \textit{COVID-19} trajectory indicators (Susceptible, Recovery, Infected People, Death). The resulting structure shows a causal relationship between the vaccine amount (parent node) and susceptible count (child node), thus confirming the technical correctness of the \textit{VacSIM} model through an external evaluation approach  \textbf{Figure 5}.

\section{Discussion:}

Researchers worldwide have been working round-the-clock to find a vaccine against \textit{SARS-CoV-2}, the virus responsible for the \textit{COVID-19} pandemic. Now that it is available, the distribution of the vaccines has numerous logistical challenges (supply chain, legal agreements, quality control and application to name a few) which might slow down the distribution process. To circumvent these challenges, a strong decision support system is needed \cite{avorn_regulatory_2020}. Reinforcement Learning (RL) based decision models \cite{kwak_deep_2021} have been built during COVID-19. In this paper, we have developed a novel distribution policy model, \textit{VacSIM} using Reinforcement Learning. We have pipelined an Actor-Critic using Kronecker-Factored Trust Region (ACKTR) model/ Deep Q-Networks (DQN) model and Contextual Bandit model in a feedforward way such that the output (Action and Rewards) of ACKTR/DQN model are fed into the Contextual Bandit model in order to provide a sensible Context comprising of actions and rewards. Contextual Bandits then optimize the policy considering demographic metrics such as the population of the State, Overall Hospital Beds, Overall ICU Beds, Ventilators and time series-based characteristics of the \textit{COVID-19} spread (susceptible population, recovery rate, death rate, total infected cases) as Context. While distributing the vaccine, identifying the part of the population who needs it the most is a challenging task, and in our case, we addressed it by the usage of the aforementioned Context. Rather than using the present-day count of infected and susceptible people, we have used SEIR-based projection, which makes our predicted policy more robust and trustworthy.
The percentage of vaccine distribution was the final output of VacSIM. From VacSIM, Maharastra, the most vulnerable state for covid-19, has received the maximum amount of vaccine, which showed the model could understand the context. Finally  we proposed two approaches to evaluate such models 1) SEIR-based Simulation-based approach, 2) Bayesian Network-based Causality driven approach. 
In simulation-based approaches, we have projected the Covid-19 cases after the vaccination. 
Our results indicate that the VacSIM  additionally reduces the 9039 cases compared to the naive-based approach. 
And from the Bayesian network, we found that vaccine distribution is most causally associated with susceptibility, which again showed that VacSIM output followed the design of model set-up and variability of context. 
Our model design, model evaluation are entirely novel. Any other resource allocation algorithm like Genetic Algorithm, Simulated annealing, Hill climbing is reward independent. A comparison of policy with this algorithm was shown in our dashboard (http://vacsim.tavlab.iiitd.edu.in:8000/). VacSIM outperformed these models when compared using SEIR based simulation approach.
.Finally, we provided open-source code will enable testing of our claims by other researchers. 

\section{Limitations and future work:}
\textit{VacSIM} may have some limitations shared by all Reinforcement Learning models,  i.e. the transparency of their learning process and the explainability of their decisions. The development of \textit{VacSIM} has been carried out while observing the pandemic in the past few months. However, the dynamic nature of pandemic may require a change in actions, thus calling for common sense working alongside artificial intelligence. A third limitation is the potential dependency between COVID-19 trajectory variables across states as these may not be completely independent especially after the lifting of lockdown. However, parametrizing these dependencies is a challenge and we reason that local context of the states is most strongly influenced by its own past values. 
As mentioned before, the availability of authentic data that is granular in terms of features covered is a major challenge in any kind of modelling work associated with the pandemic. By virtue of the same, we have given utmost priority to achieving herd immunity and decreasing the susceptible population as much as possible, given that proper data capturing of those features is available. The same is reflected through our choice of reward function in this work. For future work, if a more reliable and extensive data source is available describing the information of different cohorts of the Indian population on a more atomic level (such as information regarding the comorbidities of a certain region), then such relevant features should be taken into account while designing the reward function. Similarly, designing reward functions with features which are important for the efficient distribution of scarce resources like ventilators, drugs on the availability of extensive reliable data sources can be done through our framework depending on the observation and action space. \\
It should also be noted that while we assume certain vaccine characteristics, such as efficacy and gap between both doses, these are hyperparameters that can be changed depending on the availability of exact reliable data. To date, there does not exist a single well-proven statistic for the dose-wise efficacy of Covaxin and Covishield (vaccines available in India currently). Even the gap between both doses has been changed repeatedly by the Government, the latest being 12-16 weeks \footnote{https://www.bmj.com/content/372/bmj.n18} for Covishield. In the emergence and spread of various mutations of COVID-19 virus, and responding variants of vaccines entering the market, such assumptions have been taken to make the pipeline as generalised as possible, so that when required, different pieces of information can be plugged into our framework and adjusted to identify the efficiency for different resources and context. Given these limitations, we have attempted to carefully design the VacSIM pipeline and associated reward function by making realistic assumptions which can be helpful in testing such a framework before moving to on-ground deployment.  \\
To truly facilitate our ultimate goal of on-ground deployment of VacSIM, Plug-n-Play(PnP) pipeline can be created that gives policymakers the flexibility to provide varying input contexts, such as those collected on the district level. This platform would be helpful in testing our framework's efficiency  in a variety of settings, across many data points, observation spaces and even reward functions. In conclusion, we believe that artificial intelligence has a role to play in the optimal distribution of resources such as vaccines, syringes, drugs, personal protective equipment (PPEs), ICUs etc. The second wave of COVID-19 in India has also highlighted the importance of the same and our open source VacSIM framework is a potential step towards enabling use of artificial intelligence by researchers and policymakers.

\section{Conclusion:}
We provide a novel, open-source and extensible solution vaccine allocation problem that policymakers and researchers may refer to while making decisions. This feedforward network can be adapted and used for optimal allocation of various essential resources in varying scenarios and context as the Reinforcement Learning part of VacSIM can always be retrained to adapt to spatial and temporal variations by accommodating the latest values of variables like recovery rate and age distribution. Since there is some evidence that COVID-19 infections tend to show oscillatory patterns on shorter timescales \cite{bukhari_periodic_2020}, the key goal of our approach to inform an agile response to the problem of vaccine distribution has been fulfilled through this work. Although the present results are on five states, \textit{VacSIM} will be extended to all the states of India.

\section{Availability of data and material:} 
All the data and source code are available at \textit{https://github.com/tavlab-iiitd/VacSIM}

\section{Declarations}
\subsection{Conflicts of interest/Competing interests:} Not Applicable
\subsection{Funding:}This research did not receive any specific grant from funding agencies in the public, commercial, or not-for-profit sectors.
\section{Acknowledgements}
This work was partially supported by the Wellcome Trust/DBT India Alliance Fellowship IA/CPHE/14/1/501504 awarded to Dr. Tavpritesh Sethi and the Center for Artificial Intelligence at IIIT-Delhi. 

\bibliographystyle{Ref}
\bibliography{Vaccine.bib}

\end{document}